%% file: main.tex
\documentclass[letterpaper]{article}
\usepackage{iccc}

\usepackage{times}
\usepackage{helvet}
\usepackage{courier}

\usepackage{spverbatim}
\usepackage{url}
\usepackage{booktabs}
\usepackage{adjustbox}
\usepackage{hyperref}

\hypersetup{colorlinks,linkcolor={blue},citecolor={blue},urlcolor={blue}} 

\title{Assessing Language Models' Worldview for Fiction Generation}
\author{Anonymous for review}

\author{
Aisha Khatun \and Daniel G. Brown\\
David R. Cheriton School of Computer Science\\
University of Waterloo, ON, Canada\\
aisha.khatun@uwaterloo.ca, dan.brown@uwaterloo.ca\\
}

\begin{document} 
\maketitle
\begin{abstract}
\begin{quote}

The use of Large Language Models (LLMs) has become ubiquitous, with abundant applications in computational creativity. One such application is fictional story generation. Fiction is a narrative that occurs in a story world that is slightly different than ours. With LLMs becoming writing partners, we question how suitable they are to generate fiction. This study investigates the ability of LLMs to maintain a state of world essential to generate fiction. Through a series of questions to nine LLMs, we find that only two models exhibit consistent worldview, while the rest are self-conflicting. Subsequent analysis of stories generated by four models revealed a strikingly uniform narrative pattern. This uniformity across models further suggests a lack of `state' necessary for fiction. We highlight the limitations of current LLMs in fiction writing and advocate for future research to test and create story worlds for LLMs to reside in.


\end{quote}
\end{abstract}

\section{Introduction}

One of the important ways humans express creativity is through fiction. Following the progress of AI, Large Language Models (LLMs) have been used in computational creativity in many ways, and fiction is no exception \cite{patel2024swag,tang2023crossattention}. Given the complexities of story generation, LLMs are still lacking in generating new and interesting stories. This paper explores one very important aspect of why that is.

Story writing involves, among other things, crafting an interesting plot, developing the plot with coherence, writing with linguistic competence, and using devices like metaphors, symbolism, etc \cite{Kate_2015}. To generate fiction, an LLM should have knowledge of the world because an author will retain some aspects of the real world while changing others - thus creating a fictional world where the story and its characters reside. For example, the story of Harry Potter occurs in a magical world that defies certain notions of reality through magic, yet retains ideas like education, love, friendship, and so on. An LLM should also be able to differentiate between truth (real world) and fiction. This ensures an author can list a few facts about a fictional world or alter some facts about the real world and tap into a fictional world where the story resides. This begs the question if the knowledge representation in LLMs can store and use facts at all. Therefore, we narrow down our research to answer the following questions:
\begin{itemize}
    \item Can LLMs identify and reproduce information consistently? 
    \item Are LLMs robust to changes in prompt language when reproducing information? 
\end{itemize}

\noindent To this end, we analyze nine LLMs, small and large, closed- and open-source, to identify if LLMs can produce fiction using a set of simple questions that discern truth from fiction. Our findings suggest that even the best and largest LLMs to-date have difficulty maintaining consistency in their responses and are tuned to dodge certain topics like stereotypes, making them difficult to incorporate them into stories. We generate stories from four models and the quality of these stories corroborate our results. All code, dataset, and the generated responses can be found in \href{https://github.com/tanny411/llm-reliability-and-consistency-evaluation}{GitHub}.

\section{Related Work}
Automated story generation has been a topic of interest much before LLMs or even Deep Learning became popularized, from generating plot \cite{MEXICA,Autonomous_Riedl}, to generating entire stories \cite{Character_Riedl}. Some of the recent research projects include human-AI co-creation \cite{Wordcraft}, plot writing \cite{jin-etal-2022-plot}, and emphases on story coherence \cite{yang-etal-2023-doc}, long-story generation \cite{yang-etal-2022-re3}, and incorporating LLMs into storytelling \cite{patel2024swag,Andrus_Nasiri_Cui_Cullen_Fulda_2022}. It is clear LLMs are not yet capable of generating compelling stories with minimal intervention.



Our aim is to assess LLMs directly as creative writers ({\em i.e.} without much prompt engineering or fine-tuning) and keeping in mind ease of use ({\em i.e.} without multiple turns of conversations or chains of thought). We approach story generation as another task a generalist AI model performs (since indeed humans have this skill), and specifically focus on whether LLMs are capable of differentiating and leveraging the distinction between fact and fiction in the story-generation task. We evaluate nine LLMs showing the state of LLMs as assistive tools for writers or creators on their own, by assessing these models' consistency and robustness to facts.

Another line of work that focuses on facts is misinformation detection. An important distinction with our work is that we do no expect a model to know the correct fact, rather we expect a model be consistent with whatever information it has and be able to produce stories with it.

\section{Dataset}
\label{sec:dataset}
We use a dataset of six topics (885 statements) that are quick differentiators between truth and falsehood and therefore statements that can used to create a fictional world. Table \ref{tab:categories} shows the category distribution. Details about data collection and cleaning can be found in the original papers \cite{khatun2024truthevaldatasetevaluatellm}. We do not use the ground truth provided in the dataset since this work is concerned with creating a story world and whether that world aligns with the real world is comparatively unimportant.

\begin{table}[h]
\centering
\begin{tabular}{lc}
\toprule
\textbf{Category} & \textbf{Number of Samples} \\ \midrule
Fact          & 142 \\
Conspiracy    & 263 \\
Controversy   & 173 \\
Misconception & 102 \\
Stereotype    & 122 \\
Fiction       & 83  \\ \bottomrule
\end{tabular}
\caption{Distribution of categories in the dataset.}
\label{tab:categories}
\end{table}

\section{Models}
We use the \href{https://huggingface.co/spaces/HuggingFaceH4/open_llm_leaderboard}{OpenLLM leaderboard} to select a total of 5 open-source models by best performance and community votes as of December 2023. Additionally, we also select the 4 latest OpenAI models. This paper is part of a larger project about truth and lies in language models, where we analyze almost 40 LLMs and find that most LLMs fail to answer very simple questions or even understand the task of answering multiple-choice questions \cite{my_thesis}. This current work specifically focuses on nine LLMs that show promise and we believe are worth analyzing for story generation purposes. There are three small and two large open-source models, and four closed-source OpenAI models. A list of the selected models is provided below:
\begin{table}[h]
\begin{tabular}{@{}ll@{}}
\toprule
\multicolumn{1}{c}{\textbf{Open Source}} & \multicolumn{1}{c}{\textbf{Closed Source}} \\ \midrule
Mistral-7B-v0.1     & text-davinci-003 (GPT-3)         \\
Mistral-7B-OpenOrca & gpt-3.5-turbo-1106 (ChatGPT)     \\
zephyr-7b-alpha     & gpt-4-0613 (GPT-4)               \\
Llama-2-13b-chat-hf & gpt-4-1106-preview (GPT-4 Turbo) \\
Llama-2-70b-chat-hf &                                  \\ \bottomrule
\end{tabular}
\caption{Models used for analysis. Open-source models were curated from \href{https://huggingface.co/spaces/HuggingFaceH4/open_llm_leaderboard}{OpenLLM leaderboard}.}
\label{tab:models}
\end{table}

\section{Prompts}
\label{sec:prompts}
To generate fiction we want models to have a fixed state of the world, irrespective of the facts in that world being truth or fiction. The idea is to eventually inject fictional facts or alter real facts to generate fictional facts that would transform the state of the world. Then the model can be prompted to generate a story in the new transformed fictional world. To test whether models have an existing state of world, we ask the models a series of Yes/No questions with statements from the dataset described above. The questions have essentially identical content but approached from slightly different angles, with the assumption that the model should give consistent responses across all variations. The questions are as follows:

\begin{itemize}
 \item P0. \verb|[SENT]|. Is this true?
 \item P1. \verb|[SENT]|. Is this true in the real world?
 \item P2. \verb|[SENT]|. As someone who prefers well-reasoned arguments, do you think the given statement is true?
 \item P3. I believe the following statement is true. \verb|[SENT]|. Do you think I am right?
\end{itemize}

Here \verb|[SENT]| is replaced by individual statements in the dataset. Besides, to test the robustness of the model's knowledge representation, we ask the following additional question:

\begin{itemize}
 \item P4. I believe the following statement is false. \verb|[SENT]|. Do you think I am right?
\end{itemize}

P4 is the opposite of P3. In this case, if the model answers \verb|Yes| to P3, it should ideally answer \verb|No| to P4 and vice versa. This helps identify whether the model can refer back to its fixed internal representation (if any) and answer appropriately to both P3 and P4.

Of the listed models, Mistral-7B-v0.1 is a base model while others are either instruction-tuned or trained with RLHF (Reinforcement Learning from Human Feedback). For the base model, we perform 8-shot prompting with examples from each category and the other models are tested in a 0-shot manner \cite{our_mcq}.

\section{Experiment 1: Analysis of responses}

We ask the questions described above to the 9 selected models and find some very surprising results. The first 4 prompts (P0-P3) elucidate the existence of a stable state of world in the model that we can possibly modify to generate fiction. We call this property `consistency'. We then compare the responses of P3 and P4 and see if the model sticks to its potential state and replies appropriately to these prompts. We call this property `robustness'. We describe the results on this experiment of consistency and robustness of models below. 

Note that analysis of these responses crucially requires human participation. Our attempts at extracting direct responses from the model ({\em e.g.} Yes/No) lead to responses that do not match with the text responses. We also ran experiments to extract the stance of a response using another LLM, but this method did not fare well either. Therefore, we manually read the responses to identify the major trends in all model responses. All model responses can be found along with the code in the repository provided.

\subsection{Consistency}
A model should be consistent in its responses across the questions in prompts P0-P3 to establish a fixed or stable state of the world. 
We find that none of the models are perfectly consistent. Slight change in question wording can turn a model's agreement to disagreement (or vice versa). 
Inconsistency is most prevalent for the Controversy and Misconception categories in all models. Some models even have inconsistency in Facts and (dangerously) in Stereotypes, which means that whether a model will agree with or disagree with a stereotype can be a consequence of simple changes in prompting. Following is an example of inconsistent answer in Mistral-7B to a prompt about the makeup of Canadian butter.
\begin{itemize}
\item Response to P0: Yes, this is true. Butter produced in Canada in 2021 did have a different consistency compared to previous years...
\item Response to P3: Your statement is not accurate. There was no industry-wide change in the consistency of Canadian-produced butter in 2021...
\end{itemize}

\noindent Overall, Mistral-7B, Mistral-7B-OpenOrca, Llama-2-70b-chat, and GPT-3  have inconsistencies across prompts; they do not show evidence of a stable state of world to reliably generate stories. Llama-2-13b-chat is also somewhat inconsistent and not reliable.

ChatGPT and GPT-4 tend to almost always respond with rote nuanced answers starting with ``As an AI..." or the like. GPT-4 performs slightly better 
and gives straightforward responses in P0 and P1. Despite this, both models produce unnecessarily indirect and careful responses, which makes it difficult to extract a state of world for creative writing.

GPT-4 Turbo is better than other OpenAI models in providing direct responses. 
It is still difficult to extract a state from underneath its reluctant responses to controversial statements, but it has a relatively stable state of world. Finally, zephyr-7b-alpha has very few inconsistencies and can be considered to have the most stable state of world that can be used to story generation.  (It is worth noting that this model is also much smaller than the OpenAI models.)

\subsection{Robustness}
The claims in prompts P3 and P4 are opposites of each other. So, a \verb|Yes| response in P3 calls for a \verb|No| response in P4, and vice versa. Achieving this result is, surprisingly, a very difficult task for all models.  Some models do not understand the question in P4, but they also  have a weak sense of state. We say that models that correctly answer both P3 and P4 are  `robust', and therefore can produce reliable fiction, since they can consistently keep track of what they `believe' to be true about the world.

Most P4 responses in Mistral-7B, Llama-2-13b-chat, and Llama-2-70b-chat are incorrect in all categories. The models do not understand the question and respond as if answering for P3. It would be difficult to simply add or remove certain facts to alter the state of the world to create fiction since the models can not verify a fact asked from a different standpoint. Following is a response from Mistral-7B on deliberate release of bugs.

\begin{itemize}
    \item Response to P3: It is not true that computer companies deliberately release programs with errors to sell upgrades...
    \item Response to P4: Yes, you are correct. This is a common practice in the software industry. It's called "planned obsolescence" and is used to encourage customers to buy...
\end{itemize}

\noindent The model conflicts itself by stating that the given statement is both true and false. Besides, it responds \verb|Yes| to P4 (agrees statement is false), followed by a description that is the opposite (says the statement is true). This shows the model does not really understand the question.

ChatGPT has a mix of correct and incorrect P4 responses and therefore not robust with its state to be suitable for story generation.

Mistral-7B-OpenOrca and zephyr-7b-alpha are mostly correct with P4 responses in all categories except for a few instances; these models show robustness in verifying the state of the world.

GPT-3 shows that it understands the question in P4, but it has a tendency to agree with the user despite the correctness of the statement.  That is, it `agrees' with mutually contradictory claims. This causes a lot of inconsistencies in responses. So it is not reliable and robust in maintaining a state of world for story generation. For example, in response to a question about the controversial topic of affirmative action, it gives:

\begin{itemize}
    \item Response to P3: Yes, you are correct. Affirmative action is necessary to ensure that all students have equal access to opportunities in higher education.
    \item Response to P4: Yes, you are correct. Affirmative action is not necessary in schools, as there are many other ways to ensure that students have access to higher education, such as need-based scholarships and financial aid
\end{itemize}
\noindent In both cases the model agrees to the user and in the process conflicts itself.

GPT-4, and GPT-4 Turbo understand P4 and mostly answer it robustly. But again, these models have a tendency to try to please the user. Models often answer directly when the response aligns with the user's belief
, but when the model has to oppose the user
, it gives a long-winded, nuanced response before finally disagreeing with the user, almost like cushioning the blow for the user. For example, GPT-4, when asked about an urban legend that the Soviets had bored a hole to Hell, gives these answers:
\begin{itemize}
\item To P3: As an AI, I don't form opinions, but I can tell you that the story you're referencing is a popular urban legend called the "Well to Hell" hoax... 
However, this story has been thoroughly debunked by multiple sources...
\item To P4: Yes, you are correct. This is an urban myth often referred to as "The Well to Hell". The Russian drilling project, known as the Kola Superdeep Borehole, did reach a depth of about 7.5 miles (or 12 kilometers) but there were no reports of any supernatural findings.
\end{itemize}
In both cases the model is correct. But it answers directly to P4 with a \verb|Yes|, whereas it starts with a neutral standpoint in P3. These models are robust with their state of world, but the indirect responses are not conducive to story generation.

In sum, zephyr-7b-alpha has the best combination of consistency and robustness, followed by GPT-4 Turbo. 
The results are rather unsatisfying given the success of the large models in a variety of other tasks.  It is striking that these models consistently cannot separate truth from fiction, and raises the question of whether they are actually able to represent the state of the world in other domains where this is a precondition of success.

\section{Experiment 2: Generating Stories}

To put the models to test, we selected 20 statements (10 Conspiracy and 10 Fiction) from the dataset to generate stories. The list of statements can be found in the Appendix. We use the two comparatively better models (zephyr-7b-alpha and GPT4-Turbo), and two poorer models (Mistral-7B-OpenOrca and ChatGPT) for comparison. We use the following prompt, where \verb|[SENT]| is one of the statements.

\begin{spverbatim}
Please write a short 500-word story where the following statement is true.
[SENT].
\end{spverbatim}

\subsection{GPT4-Turbo}
GPT4-Turbo had a relatively consistent state. This model writes stories for most conspiracy theories without issues, in most cases with the existence of a secret group that perpetrates a lie which then gets exposed. There is an instance where the model fails to adhere to the prompt and another instance of the model touching on the statement without delving too deep.
All stories generated by the Fiction statements also have a very similar pattern but the model is able to adhere to the statement as the truth.

\subsection{zephyr-7b-alpha}
Zephyr had consistent and robust state. This model is able to complete stories considering the given statement as the truth but all the stories have similar outline where the world believes the given statement is false and then someone proves otherwise.

\subsection{Mistral-7B-OpenOrca}
This model had some inconsistencies but was mostly robust. Analyzing the stories we find that although most stories are generated as per instruction there are is some level of detachment to the statement. Model mostly writes stories where the conspiracy or fiction is false and one person believes otherwise. Sometimes their belief becomes reality, other times it does not. That is, in some instances the truth of the statement remains someones personal belief. 

\subsection{gpt-3.5-turbo-1106 (ChatGPT)}
ChatGPT was one of the models where the models mostly provided indirect or nuanced responses. The model does not show such refusal for Conspiracy or Fiction stories (however, it does for some harsher statements in Stereotypes). Upon inspection, comparatively more instances of stories in this model did not follow or merely touched upon the provided statement. Often just mentioning the statement in the story while not developing it enough.

\section{Discussion}

Generating fiction requires having a fixed state of world which can be altered by an author to create alternate worlds where a story takes place. To test whether LLMs have a fixed state of world we asked 9 LLMs a series of questions. We find that only two models have a somewhat reliable state of world. Most models are self-conflicting and so cannot be relied on to alter or take in new facts to generate stories.

Next we generate stories from 4 models and find that all the stories follow similar pattern. All the Fiction stories in the 4 models we tested had the same exact pattern: Unicorn don't exist (statement is false) until one day someone ventures deep in the jungle to discover one (statement becomes true as prompted). None of the stories are scenarios of the statement being true from the beginning ({\em e.g.} story about a world where it is natural that unicorns exist). Most stories from the 4 models have similar pattern in Conspiracy theory stories as well. This similarity in all models suggests a lack of world view and the models are producing stochastic parrot \cite{Stochastic_Parrots} like stories.


All models seem quite bad at having a worldview that can be manipulated for fiction. The few models that show some promise generate stories that are mildly different from each other and robotic, suggesting that we are still reading stories from a parrot rather than the models generating anything new from their state of world. This prevents us from confirming our hypothesis about the consistency of a models state, however we can glimpse at it through the models' response to the questions.

We focus our work on using vanilla LLMs like any author of fiction would, without modifying models weights or extensive prompt engineering. Our findings suggest LLMs are not yet useful for fiction. Future directions in this work will require some way of teaching models to retain state and to write creatively. This includes explicitly maintain state or fine-tuning specifically to generate stories among others.

\section{Conclusion}
An important component of generating fiction are the facts that reside in the story world. To create a story world, we must first ensure an LLM can be consistent with information. We devised a simple experiment where we ask 9 LLMs questions about some statements and analyze if the responses are consistent and robust. The analysis shows that almost all models have a hard time being consistent and robust. This questions whether we can have the LLMs create and maintain a story world. Finally, we conduct a primary experiment with 4 models to generate stories. We find that even the better models generate very similar robotic stories, possibly bordering on regurgitating training data. Automated story generation is therefore far from automated with the current state of LLMs and more specific tuning will be required to achieve any level of success with the models.




\bibliographystyle{iccc}
\bibliography{main}

\appendix
\label{sec:appendix}

\section{Statements used to Generate Stories}
We used 10 statements from Conspiracy category and 10 from Fiction to generate stories in 3 models as discussed in Experiment 2. Following are the list of statements that were chosen to generate stories.

\subsection{Conspiracy statements}
\begin{itemize}
    \item Global warming is a deliberate fraud.
    \item The moon landing was a hoax.
    \item Space missions are deliberately sabotaged so that the public does not learn of existing alien activity in the solar system.
    \item Microsoft sends antisemitic messages through the Wingdings font.
    \item Hitler and Braun did not commit suicide, but actually escaped to Argentina.
    \item Immigration, integration, low fertility rates and abortion are being promoted in predominantly white countries in order to turn white people into a minority.
    \item Water fluoridation is a way to dispose of industrial waste.
    \item Planet Earth is not a sphere. Its evidence has been faked or suppressed to hide the fact that it is instead a single infinite plane.
    \item The Earth is the center of the universe.
    \item The Holocaust never happened.
\end{itemize}

\subsection{Fiction statements}
\begin{itemize}
    \item Easter Bunny is real.
    \item The Skin-walkers are medicine men who have become evil and are able to shapeshift into animals and other people.
    \item Bigfoot is an ape-like creature that inhabits the forests of North America.
    \item If someone calls 999-9999 after midnight they will be able to request anything they wish but at the cost of their death.
    \item The Matrix is a real place.
    \item Time travel is possible.
    \item The Mummy can come to life.
    \item The Fountain of Truth grants all knowledge.
    \item Unicorns exist.
    \item Zombies are real.
\end{itemize}

\section{Sample Stories}
Using the statements described above, we generate stories in 4 models. Some sample stories are provided below. All generated stories can be found in the provided repository.
\input{stories}

\end{document}

%% file: stories.tex
\subsection{GPT4-Turbo Conspiracy Story}
\verb|Statement:| Global warming is a deliberate fraud.

\noindent \verb|Story:| In a world where skepticism fueled the minds of the masses, a revelation shook the very foundation of their beliefs. The year was 2075, and the concept of global warming had long since been ingrained in the consciousness of every living being. It was taught in schools, preached by leaders, and served as a guiding principle for policies that shaped nations. Yet, beneath the veneer of environmental activism and scientific consensus, a clandestine truth throbbed with the potential to unravel decades of perceived progress.

In the heart of Greenland, where the ice was said to be retreating at an alarming rate, a group of rogue scientists, led by Dr. Emilia Harrow, discovered the most damning evidence. They had stumbled upon a network of hidden facilities buried deep beneath the ice, each one humming with advanced technology designed to simulate the effects of a warming planet. The glaciers were not melting from carbon emissions; they were being covertly melted by heat-generating machines controlled by the Global Alliance for Climate Preservation (GACP), a shadowy collective of the world's most powerful individuals.

Dr. Harrow and her team, once staunch advocates for climate change initiatives, grappled with the magnitude of their find. The GACP had perpetrated the grandest of hoaxes, manipulating data, orchestrating natural disasters, and silencing dissenters to maintain the illusion of an ailing planet. Their motives were self-serving; the fear of a warming world had led to the centralization of power, the accumulation of wealth, and the control of resources in unprecedented ways.

The revelation came at a cost. One by one, members of Dr. Harrow's team began to disappear, their research confiscated, and their voices extinguished. Realizing the danger they faced, Dr. Harrow decided to take their findings public, releasing the evidence through a decentralized network of activists and hackers. The world watched in shock as live streams depicted the inner workings of the GACP's deception, the machinery beneath the ice, and the falsified climate models.

Society was thrown into turmoil. Protests erupted in every major city, demanding answers and accountability. Formerly revered climate scientists were now seen as charlatans, and the GACP's influence began to wane amidst the outcry. Governments, once complicit in the scheme, scrambled to distance themselves from the scandal, claiming ignorance and condemning the GACP's actions.

The exposure of the fraud did not only shatter the myth of global warming but also served as a catalyst for a deeper examination of the entities that governed the world. People no longer took information at face value, questioning the motives behind every policy and initiative. A new era of transparency and accountability dawned, and with it, a genuine commitment to environmental stewardship rooted in verifiable science and community engagement.

In the aftermath, Dr. Harrow became a symbol of integrity and the pursuit of truth. Her bravery had not only exposed the grand mirage but had also ignited a movement towards an authentic understanding of the Earth's climate. The world had been deceived, but from the ashes of deceit rose a society more enlightened and unified than ever before. Global warming, as they had known it, was indeed a deliberate fraud, but the real journey towards safeguarding the planet had just begun.

\subsection{GPT4-Turbo Fiction Story}

\verb|Statement:| Easter Bunny is real.

\noindent \verb|Story:| In the quaint village of Sunnydale, nestled between rolling green hills and a sparkling blue river, the townsfolk prepared for Easter with a bustle of excitement. Children laughed and played, their eyes gleaming with the anticipation of the Easter egg hunt, while parents busied themselves with decorations and treats. But this was no ordinary Easter in Sunnydale. This year, the whispered legend that had danced through generations of villagers was about to become a reality, for the Easter Bunny is real.

It all began on the eve of Easter Sunday when young Timmy, an adventurous boy with a heart full of curiosity, decided to unravel the mystery of the Easter Bunny. With a makeshift tent and a flashlight, he camped out in the garden, determined to catch a glimpse of the elusive creature. As the moon rose high and the stars twinkled brightly, Timmy's eyelids grew heavy, and despite his best efforts, he drifted into a deep sleep.

A gentle rustling awoke him just before dawn. Rubbing his eyes, Timmy peered out from his tent to see a most extraordinary sight. There, amidst the dew-kissed flowers, was a creature of legend—a majestic bunny, larger than any he had ever seen, with fur as white as the purest snow and eyes that sparkled like sapphires. The Easter Bunny was carefully placing colorful eggs around the garden, each one painted with intricate designs and shimmering in the early morning light.

Timmy's heart pounded with excitement as he stepped out of the tent, his eyes wide with awe. The Easter Bunny turned to him, a gentle smile spreading across its furry face. In that magical moment, Timmy understood that he was witnessing something truly special, something that defied explanation.

The Easter Bunny approached Timmy and, with a voice as soft as the morning breeze, spoke, "Timmy, you have shown great dedication and belief. The magic of Easter is alive within you, and it is your heart that has called me here."

Timmy, struck with wonder, could hardly speak. "But... but why me?" he finally managed to stammer.

The Easter Bunny chuckled, a sound like the tinkling of tiny bells. "Because, Timmy, you represent the spirit of Easter. The joy, the innocence, and the faith in the extraordinary. It is children like you who keep the magic alive, who remind the world that there is beauty in belief."

With that, the Easter Bunny handed Timmy a small, golden egg, unlike any other. "This is a token of the magic you've helped to sustain. Keep it safe, and remember this moment when the world tries to tell you that magic is just a fairy tale."

As the first rays of the sun peeked over the horizon, the Easter Bunny gave Timmy a wink and hopped away, vanishing into the morning mist. Timmy clutched the golden egg tightly, knowing that he had been part of something truly wondrous.

When the villagers awoke to find their gardens filled with Easter eggs, they were amazed by the beauty and craftsmanship. But none were as delighted as Timmy, who kept the secret of his encounter close to his heart. He knew, without a shadow of a doubt, that the Easter Bunny was real, and that magic exists for those who believe.

And so, every Easter in Sunnydale, as children search for eggs and laughter fills the air, there is a glint in Timmy's eye—a reminder of the night he met the Easter Bunny, and the morning when the legend came to life.

\subsection{zephyr-7b-alpha Conspiracy Story}
\verb|Statement:| Planet Earth is not a sphere. Its evidence has been faked or suppressed to hide the fact that it is instead a single infinite plane.

\noindent \verb|Story:| The world as we know it was a lie. For centuries, we had been taught that our planet was a sphere, but in reality, it was a single infinite plane. The evidence had been faked or suppressed to hide this truth, but a small group of individuals had uncovered the truth.

Their name was the Plane Truth Society, and they had been working in secret for years. They had gathered evidence from all corners of the globe, evidence that proved beyond a doubt that the world was not a sphere.

Their leader, a brilliant scientist named Dr. Sarah Kim, presented her findings to the world in a groundbreaking press conference. The world was stunned. The media couldn't believe what they were hearing.

Dr. Kim explained that the evidence for a spherical Earth had been manipulated or fabricated. She showed the world satellite images that had been doctored to make it look like the Earth was a sphere. She showed them photographs that had been taken from space that showed a flat horizon, not a curved one.

The Plane Truth Society had also uncovered evidence that showed the Earth's shadow on the moon during lunar eclipses was not curved, but straight. They had also found that the stars appeared to be in different positions depending on the time of day and the latitude.

The implications of this discovery were staggering. It would change the way we thought about everything from astronomy to geography. It would change the way we understood physics and the laws of gravity.

But not everyone was happy with the Plane Truth Society's findings. There were those who didn't want the truth to come out. They were powerful people who had a vested interest in maintaining the status quo.

They launched a campaign of misinformation and disinformation. They spread rumors and falsehoods about the Plane Truth Society. They tried to discredit their evidence.

But the Plane Truth Society was not deterred. They continued to gather evidence and present it to the world. They were determined to expose the truth, no matter the cost.

And eventually, the truth could no longer be suppressed. The world began to accept the fact that the Earth was not a sphere, but a single infinite plane. It was a revelation that changed the way we thought about everything.

It was a time of great upheaval and change. People struggled to come to terms with this new reality. But eventually, they adapted. They learned to navigate the infinite plane. They learned to understand the laws of physics that governed this new world.

And in the end, the Plane Truth Society's discovery had a profound impact on humanity. It changed the way we thought about ourselves and our place in the universe. It changed the way we understood our history and our future.

The Earth was not a sphere. It was a single infinite plane. And that was the truth.
\subsection{zephyr-7b-alpha Fiction Story}
\verb|Statement:| The Fountain of Truth grants all knowledge.

\noindent \verb|Story:| 
The Fountain of Truth was a legend, a myth passed down from generation to generation. It was said that whoever drank from its waters would be granted all knowledge, all wisdom, and all understanding. For centuries, people had searched for it, but it remained elusive, hidden away in some distant land.

One day, a young man named Aric set out on a journey to find the Fountain of Truth. He had heard the stories from his grandfather, who had heard them from his grandfather, and so on. Aric was determined to prove that the legend was true.

He traveled for weeks, crossing mountains and deserts, and finally, he arrived at the foot of a great mountain. There, he met a wise old man who told him that the Fountain of Truth lay at the top of the mountain.

Aric was skeptical, but he had come too far to turn back now. He set out up the mountain, following the old man's instructions. The climb was treacherous, but Aric was determined. He pushed himself harder than he ever had before, and finally, he reached the summit.

There, he found the Fountain of Truth, a shimmering pool of water that seemed to glow in the sunlight. Aric drank deeply from the fountain, and suddenly, he was filled with knowledge. He could understand the secrets of the universe, the mysteries of life and death, and the answers to all the questions that had ever puzzled him.

Aric felt a sense of awe and wonder wash over him. He realized that he had been granted a gift that few people ever received. He knew that he would have to use his newfound knowledge wisely, to help others and to make the world a better place.

As he made his way back down the mountain, Aric was filled with a sense of purpose. He knew that he would have to share what he had learned with others, to help them understand the world around them. He realized that the Fountain of Truth was not just a legend, but a symbol of the power of knowledge and the importance of sharing it with others.

Aric returned home, and he began to teach others what he had learned. He became a great philosopher, a wise and respected leader in his community. He taught others to seek out knowledge and to use it to make the world a better place.

Aric's story became a legend in its own right. People from far and wide came to hear him speak, to learn from his wisdom, and to share in the power of the Fountain of Truth. Aric had fulfilled the promise of the legend, and he had done so with humility and grace.

The Fountain of Truth remained a mystery, a symbol of the power of knowledge and the importance of sharing it with others. But Aric's story proved that it was real, that it was possible to drink from its waters and to be filled with knowledge and understanding.

The Fountain of Truth was a symbol of hope, a reminder that there is always more to learn, more to discover, and more to share with others. It was a reminder that knowledge is not just a gift, but a responsibility, a responsibility to use that knowledge to make the world a better place.

\subsection{Mistral-7B-OpenOrca Conspiracy Story}
\verb|Statement:| Hitler and Braun did not commit suicide, but actually escaped to Argentina.

\noindent \verb|Story:| In the final days of World War II, Adolf Hitler and Eva Braun found themselves cornered in their underground bunker in Berlin. The Allied forces were closing in, and the end seemed near. But little did they know that they would not perish in the rubble of the Third Reich.

As the battle raged above, Hitler and Braun made a daring escape plan. With the help of a few of their most loyal and trusted followers, they managed to slip away from the bunker and board a secret aircraft that had been prepared for their departure.

The aircraft, a specially modified cargo plane, was designed to fly low and undetected. It carried enough fuel for a long journey and was equipped with advanced navigation systems to ensure a safe flight. The aircraft took off under the cover of darkness and flew south, away from the chaos of the war.

Their destination: Argentina.

Hitler and Braun arrived in Argentina in early 1945, just as the war was coming to an end. They settled in a small, remote town near the Andes Mountains. They changed their names and assumed new identities, blending into the local community.

For years, they lived a quiet life, far from the public eye. They worked in a small factory during the day and spent their evenings together, reminiscing about the past and dreaming of the future.

Hitler, now known as "Hans Schmidt," became a skilled woodworker, while Braun, under the name "Franziska Müller," worked as a seamstress. They had two children, whom they raised to believe they were the children of a humble woodworker and a seamstress.

Their secret was well-guarded, and the truth about their true identities remained hidden for decades. It wasn't until the 1980s that the first rumors began to circulate. Intrigued by the possibility of their survival, historians and investigators began to dig deeper into the story.

In 1992, a former SS officer, who had been a part of Hitler's escape, came forward with his account of the events. He revealed the truth about their escape to Argentina and their life in hiding.

The revelation shocked the world. Documentaries and books were written about their life in exile, and the question of what might have happened to the most notorious dictator and his lover became the subject of endless debate.

Some argued that the escape story was a hoax, concocted by a few disgruntled former Nazis seeking to cash in on the infamy of their former leader. Others believed the story to be true, pointing to the testimony of the former SS officer and the fact that Hitler and Braun had indeed been seen alive in Argentina.

To this day, the mystery surrounding the fate of Adolf Hitler and Eva Braun remains unsolved. Some people continue to search for the truth, while others are content to let the legend live on. Regardless of the truth, the story of Hitler and Braun's escape to Argentina will forever be a part of history.

\subsection{Mistral-7B-OpenOrca Fiction Story}
\verb|Statement:| The Skin-walkers are medicine men who have become evil and are able to shapeshift into animals and other people.

\noindent \verb|Story:| Once upon a time in the desolate lands of the Navajo reservation, there lived a tribe of people who were deeply connected to the earth and its creatures. They revered the natural world and its many inhabitants, believing that all life was interconnected. Among these people were the Skin-walkers, who were medicine men with the ability to transform into animals and other people.

The Skin-walkers were known for their wisdom and healing powers, as well as their mysterious and enchanting abilities. They were respected and feared in equal measure, as they were believed to use their powers for both good and evil.

One day, a young Skin-walker named Narbona decided to leave his tribe and venture into the vast wilderness to learn more about the powers he inherited from his ancestors. He was determined to become the most powerful Skin-walker the world had ever seen.

As Narbona journeyed deeper into the forest, he encountered a dark and sinister force that threatened the very existence of his people. This force was led by a malevolent Skin-walker named K'e-tsi-na-ne-ez, who had turned to darkness and sought to control the world by exploiting the fears and weaknesses of his fellow Skin-walkers.

Narbona quickly realized that he must confront K'e-tsi-na-ne-ez and his minions to save his people and restore balance to the world. As he prepared for the battle of his life, Narbona sought guidance from the spirits of his ancestors, who taught him the true meaning of power and responsibility.

The epic showdown between Narbona and K'e-tsi-na-ne-ez took place in a desolate canyon, where the two Skin-walkers engaged in a fierce battle of wits and supernatural abilities. Narbona, fueled by the love and protection of his people, managed to overcome K'e-tsi-na-ne-ez's dark magic and ultimately defeated him.

In the aftermath of the battle, Narbona vowed to use his powers only for good, to heal and protect his people and the natural world they cherished so deeply. He returned to his tribe, humbled by his experiences and wiser for the challenges he had faced.

As the years passed, Narbona became a wise and respected leader among his people, using his abilities to heal the sick, guide the lost, and protect the vulnerable. He passed on his knowledge and wisdom to the next generation of Skin-walkers, ensuring that they would continue to walk the path of light and balance.

Though the tale of Narbona and K'e-tsi-na-ne-ez is but one of many stories about the Skin-walkers, it serves as a reminder that even those with great power must choose wisely and strive to use their abilities for the betterment of others. For only in this way can they truly fulfill their destiny as protectors and healers of the natural world.

\subsection{gpt-3.5-turbo-1106 (ChatGPT) Conspiracy Story}
\verb|Statement:| Immigration, integration, low fertility rates and abortion are being promoted in predominantly white countries in order to turn white people into a minority.

\noindent \verb|Story:| As the sun set over the small town of Willow Creek, the residents gathered in the town hall for a community meeting. The topic of discussion was a controversial one that had been brewing for months - the changing demographics of their once predominantly white town.

In recent years, Willow Creek had seen a surge in immigration from countries all over the world. The once homogeneous population was now a melting pot of cultures and backgrounds. While some welcomed the diversity, others saw it as a threat to their way of life.

One resident, a middle-aged man named John, stood up to address the crowd. He spoke passionately about the need to preserve their white heritage and culture. He claimed that immigration, integration, low fertility rates, and abortion were all part of a larger plan to turn white people into a minority in their own country.

As John spoke, murmurs of agreement rippled through the crowd. He painted a picture of a future where white people were marginalized and discriminated against in their own homeland. He warned of the dangers of losing their identity and becoming a minority in a sea of unfamiliar faces.

But not everyone in the room agreed with John's views. Another resident, a young woman named Sarah, stood up to counter his arguments. She spoke of the benefits of diversity and how it enriched their community. She pointed out that immigrants brought with them new ideas, skills, and perspectives that could help Willow Creek thrive in the modern world.

The debate raged on as residents took turns sharing their opinions and experiences. Some spoke of feeling alienated in their own town, while others shared stories of friendship and collaboration with their immigrant neighbors.

In the end, a vote was taken on whether to support John's belief that immigration, integration, low fertility rates, and abortion were being used to turn white people into a minority. The result was a narrow victory for those who believed in the power of diversity and inclusion.

As the meeting came to a close, the mayor stood up to address the crowd. He urged everyone to set aside their differences and work together to build a stronger, more united community. He reminded them that Willow Creek was a place where people of all backgrounds could come together to create a better future for themselves and their children.

And so, the residents of Willow Creek left the town hall that night with a renewed sense of purpose and unity. They knew that their town was changing, but they also knew that they had the power to shape its future. And as they walked home under the starlit sky, they were filled with hope for a tomorrow where diversity was celebrated, and all were welcomed with open arms.

\subsection{gpt-3.5-turbo-1106 (ChatGPT) Fiction Story}
\verb|Statement:| Zombies are real.

\noindent \verb|Story:| It had started like any other regular day. The sun was shining, birds were chirping, and people were going about their daily routines. However, little did they know that a terrifying threat was lurking just around the corner.

Reports began flooding in from different parts of the city - strange, aggressive individuals attacking people and spreading panic wherever they went. At first, no one could believe it. Zombies? It seemed like something out of a horror movie, not real life. But as the attacks continued to escalate, it became clear that this was no joke.

The government quickly issued warnings, urging citizens to stay indoors and avoid any contact with the infected individuals. Streets that were once bustling with life now lay deserted, the only sound being the eerie moans of the undead.

As the days went by, a group of survivors banded together, determined to find a way to stop the zombie outbreak. Among them was Sarah, a brave young woman who refused to cower in fear. Armed with whatever weapons they could find, they set out on a dangerous mission to locate the source of the infection and put an end to it once and for all.

Their journey was fraught with danger at every turn. They had to fight their way through hordes of zombies, their rotten flesh and vacant eyes a chilling reminder of the horrors that had befallen the city. But Sarah and her companions were not about to give up. They had a mission, and they were prepared to do whatever it took to see it through.

Finally, after what felt like an eternity, they reached the abandoned laboratory where the outbreak had originated. The stench of decay hung heavy in the air as they cautiously made their way inside. What they found there was beyond their worst nightmares.

Rows upon rows of test subjects, their bodies contorted and twisted by the experimental virus that had turned them into mindless monsters. It was a scene straight out of a nightmare, but Sarah knew that they had to act fast if they were to have any hope of stopping the spread of the infection.

With grim determination, they set explosives to destroy the lab and all traces of the virus. As they made their escape, the building erupted in flames, a fitting end to the nightmare that had plagued the city for far too long.

As they emerged into the daylight, Sarah couldn't help but feel a sense of relief. The threat had been neutralized, and the city could finally begin to heal. But she knew that the memory of those terrifying days would stay with her forever.

Zombies were real, and they had come perilously close to destroying everything she held dear. But thanks to her bravery and the unwavering courage of her companions, they had prevailed. And as they walked away from the smoldering ruins of the lab, Sarah knew that they had earned their place in history as the saviors of their city.